\documentclass{amia}

\usepackage{float}
\usepackage{amsmath}
\usepackage{amssymb}
\usepackage{graphicx}  
\usepackage{hyperref}
\usepackage{adjustbox}
\usepackage{etoolbox}
\usepackage{siunitx}
\usepackage{booktabs}
\usepackage{array}
\usepackage{multirow} 
\usepackage{subcaption}
\usepackage[parfill]{parskip}

\begin{document}

\title{ProtoBERT-LoRA: Parameter-Efficient Prototypical Finetuning for Immunotherapy Study Identification}

\author{Shijia Zhang, MBI$^1$;Xiyu Ding, MS$^1$;Kai Ding, PhD$^2$;Jacob Zhang, PhD$^2$;Kevin Galinsky, PhD$^2$;Mengrui Wang, MS$^3$;Ryan P.Mayers, MS$^4$;Zheyu Wang, PhD$^1$;Hadi Kharrazi, PhD$^{1,5}$}

\institutes{
    $^1$ Johns Hopkins University School of Medicine, Baltimore, MD; $^2$ Takeda Pharmaceuticals, Cambridge, MA; $^3$ Boston University Graduate School of Arts and Science, Boston, MA; $^4$ University of Maryland School of Medicine, Baltimore, MD; $^5$ Johns Hopkins University Bloomberg School of Public Health, Baltimore, MD}

\maketitle

\section*{Abstract}

\textit{Identifying immune checkpoint inhibitor (ICI) studies in genomic repositories like Gene Expression Omnibus (GEO) is vital for cancer research yet remains challenging due to semantic ambiguity, extreme class imbalance, and limited labeled data in low-resource settings. We present ProtoBERT-LoRA, a hybrid framework that combines PubMedBERT with prototypical networks and Low-Rank Adaptation (LoRA) for efficient fine-tuning. The model enforces class-separable embeddings via episodic prototype training while preserving biomedical domain knowledge. Our dataset was divided as: Training (20 positive, 20 negative), Prototype Set (10 positive, 10 negative), Validation (20 positive, 200 negative), and Test (71 positive, 765 negative). Evaluated on test dataset, ProtoBERT-LoRA achieved F1-score of 0.624 (precision: 0.481, recall: 0.887), outperforming the rule-based system, machine learning baselines and finetuned PubMedBERT. Application to 44,287 unlabeled studies reduced manual review efforts by 82\%. Ablation studies confirmed that combining prototypes with LoRA improved performance by 29\% over stand-alone LoRA.}

\section*{Introduction}

The treatment of immune checkpoint inhibitors (ICI) has revolutionized cancer treatment by blocking interactions between checkpoint proteins and their T cell partners, harnessing the immune system of the body to target tumors{\cite{he_immune_2020}}. Currently approved therapies target programmed cell death protein 1 (PD-1) and its ligand, programmed death ligand 1 (PD-L1), or cytotoxic T lymphocyte protein 4 (CTLA-4). As of 2022, 9 ICIs were approved for different types of cancers{\cite{zhang_cutaneous_2023}}. The eligible cancer patient population for ICI treatment increased from 1.54\% in 2011 to 43.6\% in 2020{\cite{jamal_immune-related_2020}}. In ICI research, multiomics data facilitate the early identification of therapeutic targets and biomarkers by providing comprehensive system-level insights into disease mechanisms and treatment responses during early drug development{\cite{noauthor_multi-omics_nodate}}. Despite the clinical promise of ICIs in cancer treatment and pathogenesis studies, identifying immunotherapy-related reports in large genomic repositories such as the Gene Expression Omnibus (GEO) remains a challenge. These databases contain millions of studies, but the number of studies\cite{al-danakh_immune_2022,kovacs_transcriptomic_2022} specifically focused on ICI treatment with relevant pre- and post-therapeutic genetic data remains limited. Retrieving such rare data is essential for accelerating biomarker discovery. 

To ensure accurate curation, studies must explicitly involve ICI treatment, with -omics data generated from cancer samples directly exposed to checkpoint inhibitors, ideally within a pre- vs. post-treatment experimental framework. Importantly, immune-related does not equate to ICI-exposed, and only datasets reflecting actual therapeutic intervention should be included. The conventional approach to curating multiomics data in research databases often relies on manual searches of repositories like GEO, supplemented by rule-based keyword matching systems (e.g., filtering studies containing terms like ``anti-PD-1" or ``CTLA-4")\cite{li_citsa_2023} from study abstracts. Although these systems automate initial screening, they are rigid and unable to resolve semantic ambiguity due to a lack of contextual understanding{\cite{bryant_artificial_2024}}. For example, studies that describe biological mechanisms such as "PD-L1 expression in the tumor microenvironment" can potentially be misclassified as related to immunotherapy despite the lack of therapeutic intent{\cite{kim_differential_2021}}. Similarly, research investigating the physiological role of CTLA-4 in immune homeostasis (e.g. maternal-fetal tolerance) could be classified as cancer-focused due to keyword overlap{\cite{wing_ctla-4_2008}}. In addition, if a study references other works containing these keywords, the inherent rigidity of the rule-based system may contribute to an increased rate of false positive classifications when curating research databases. For example, a study that briefly cites an article mentioning "PD-1" might be misclassified as an ICI-focused study even if it does not primarily address ICI treatments.

Advances in deep learning and natural language processing (NLP) have improved the understanding of medical text,  addressing the rigidity of conventional rule-based methods. The advent of pre-trained language models (PLMs) such as Bidirectional Encoder Representations Transformers (BERT) is capable of capturing complex semantic relationships in understanding text data based on the context{\cite{devlin_bert_2019}}. These models use self-attention mechanisms\cite{vaswani_attention_2023} to capture contextual relationships between words and were pre-trained on a vast amount of data to improve comprehension. In particular, PubMedBERT\cite{gu_domain-specific_2022} was trained from scratch on large-scale biomedical domain-specific PubMed corpora, providing a strong foundation for understanding the biomedical context.

Fine-tuning PLMs for specialized tasks such as identifying immunotherapy studies in genomic databases remains challenging in low-resource settings. Conventional fine-tuning methods, which reply on fully parametric updates, where predictions $P\left(y \mid x_i\right)$ are computed through dense layers, are prone to overfitting when training data are scarce{\cite{wu_noisytune_2022,dodge_fine-tuning_2020}}. This is exacerbated in immunotherapy studies, where labeled examples are rare \cite{al-danakh_immune_2022} and the class imbalance is extreme. 
Prototypical networks\cite{snell_prototypical_2017} address these limitations through a non-parametric approach: class prototypes are constructed as averaged embeddings of support examples, and queries are classified by their distance to these prototypes rather than learned decision boundaries. Using instance-level semantic similarities, these networks enforce a structured latent space with strong inductive biases\cite{jin_prototypical_2022}--inter-class separation and intra-class compactness-- that counteract overfitting{\cite{vinyals_matching_2017}}. 
Moreover, the episodic training paradigm inherent to prototypical networks, augments limited data by introducing combinatorial diversity through random support set selection. 
This structure is particularly suitable for the identification of immunotherapy studies, where ICI-exposed cases are limited. By learning class prototypes in the embedding space, these models classify new instances based on similarity to representative class embeddings, reducing dependence on memorizing sparse training signals.
Recent work also demonstrates that the integration of prototypical frameworks with PLM significantly improves robustness in low-resource scenarios{\cite{jin_prototypical_2022}}.

Directly fine-tuning all PLM parameters to adapt prototypical framework may risk catastrophic forgetting\cite{kirkpatrick_overcoming_2017} that erases biomedical knowledge encoded from pretraining, especially when training on a very small dataset. To avoid this, we explored the integration of Low-Rank Adaptation (LoRA){\cite{hu_lora_2021}}, which updates only low-rank decomposed matrices of PLM weights and freezes pre-trained PubMedBERT parameters to retain its biomedical literature domain-specific knowledge in our framework; LoRA helps the model maintain its ability to understand the context of immunotherapy-related terms, while also allowing it to refine features specific to the task. 

We propose a hybrid framework ProtoBERT-LoRA that integrates prototypical networks with parameter-efficient fine-tuning to identify rare immunotherapy studies in genomic databases. Unlike the previous prototypical fine-tuning framework, designed for general NLP tasks{\cite{jin_prototypical_2022}}, this is the first work that combines non-parametric prototype learning with parametric adaptation, explicitly optimized for biomedical text mining to facilitate cancer research. 

\subsection*{Background and Related Works}
\subsubsection*{Biomedical Literature Pre-trained Models}
Domain-specific PLMs are foundational for biomedical text analysis, but their effectiveness hinges on pretraining domain alignment. PubMedBERT, trained from scratch on 14 million PubMed abstracts and 3 million full-text articles, differs from models such as BioBERT \cite{lee_biobert_2020} and BioClinicalBERT \cite{alsentzer_publicly_2019} in both the pretraining strategy and domain specificity. Unlike BioBERT, which initialized from general-domain BERT weights before continued pretraining on biomedical text, PubMedBERT avoids inheriting non-scientific linguistic biases by training exclusively on biomedical corpora from PubMed. This results in better performance on domain-specific tasks that require precise biomedical terminology. BioClinicalBERT, while powerful for clinical narratives (e.g., EHR notes), is optimized for patient-centered language rather than the structured, hypothesis-driven discourse of research publications (e.g., GEO study abstracts). For our experiments, PubMedBERT’s domain-specific pre-training on biomedicine literature ensures a robust representation of academic biomedical language, critical to identifying immunotherapy studies in genomic databases.

\subsubsection*{Prototypical Learning}
Prototypical networks were originally developed for few-shot classification of unseen classes in computer vision. In each training episode, the model randomly selects a small number of examples from each class as support sets and computes their average embedding to form a class prototype. It then randomly selects additional examples as query sets and classifies each query based on its proximity to these prototypes. The training process optimizes the model by clustering query embedding close to the prototype of its own class while pushing it away from those of other classes. At inference time, prototypes computed from a limited set of labeled examples are used to classify new instances by comparing their embeddings to these class prototypes. Unlike classical metric-learning approaches such as Siamese networks \cite{Koch2015SiameseNN} and triplet networks {\cite{schroff_facenet_2015}}, which rely on pairwise comparisons between instances, prototypical networks aggregate class information into prototypes, reducing the dependency on limited instance pairs often encountered in low-resource settings.

\subsubsection*{Low-Resource Fine-Tuning}

Due to the high cost of obtaining high-quality labeled data, studies have explored strategies to fine-tune pre-trained transformer encoder models under low-resource conditions. For example, active learning approaches \cite{grießhaber2020finetuningbertlowresourcenatural} and methods based on epistemic neural networks\cite{osband2023finetuninglanguagemodelsepistemic} prioritize examples that maximize the gain of model knowledge. However, these methods cannot avoid introducing new data or complex model architectures with higher computational costs. The prototypical fine-tuning framework proposed by Jin et al. \cite{jin_prototypical_2022} combines PLMs with prototypical networks to dynamically adjust model capacity based on the size of the labeled data, achieving robust performance in general NLP tasks even with limited examples labeled. However, no studies have yet applied this framework within the biomedical domain, and its performance in medical text mining is largely unexplored.

\begin{figure}[H]  
\centering
\noindent\includegraphics[width=0.8\textwidth,keepaspectratio]{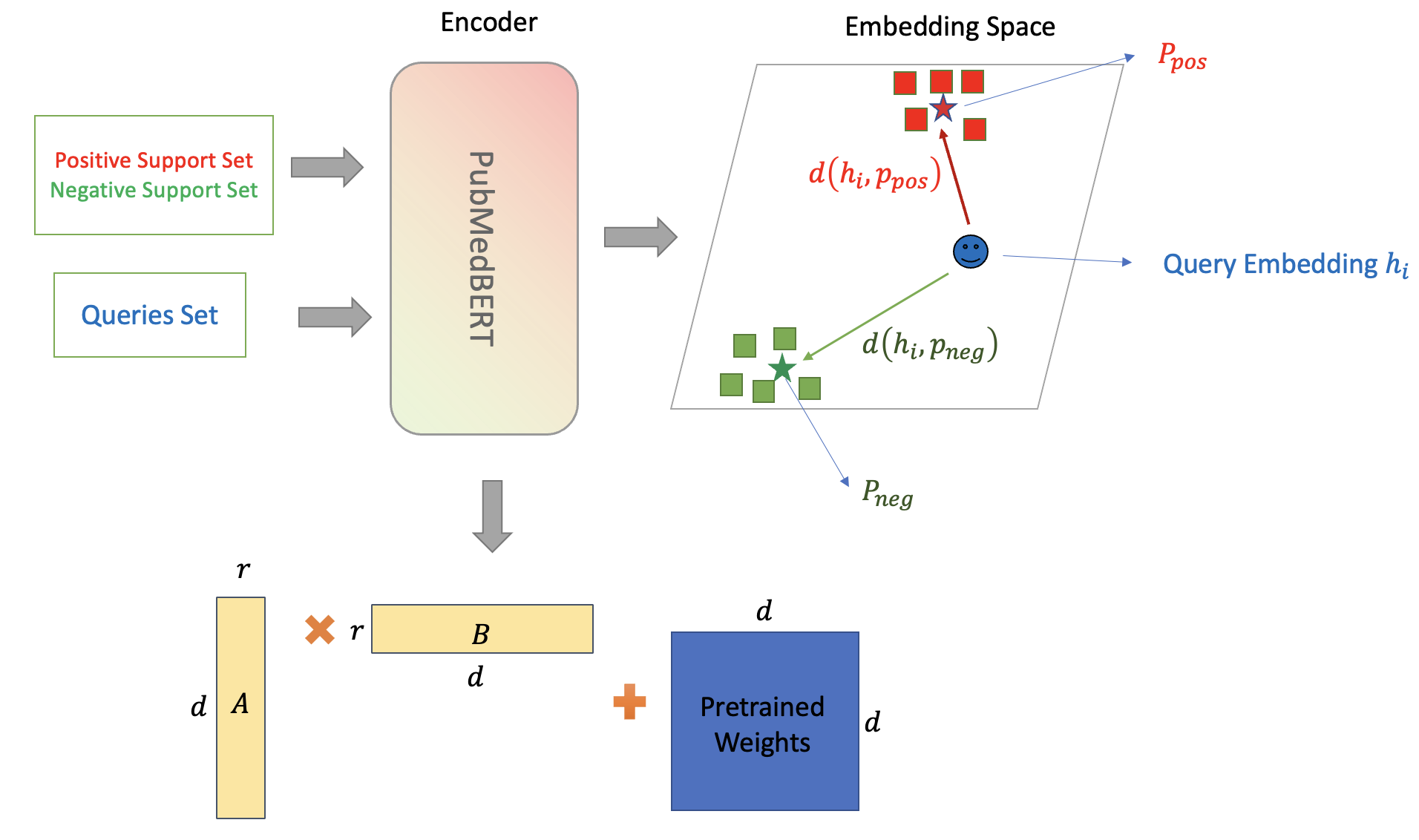}
\caption{This graph illustrates how the prototypical fine-tuning framework operates in a two-class, five-shot training scenario for each query. In the figure, $p_{pos}$ and $P_{neg}$ represent the prototypes for the ICI treatment and non-ICI classes, respectively. During training, only the $A$ and $B$ matrices are updated as part of the LoRA adaptation process.}
\label{fig:figure2}
\end{figure}

\section*{Methods}
\subsection*{Problem Formulation}
We formulate the task of identifying immunotherapy studies as a low-resource binary classification problem, where each study $x_i$ is assigned to one of two classes: $C_{pos}$ for studies involving ICI treatments and $C_{neg}$ for non-ICI studies. We adopt a meta-learning paradigm and fine-tune our framework using episodic training. In each episode, a support set is created by randomly sampling $N_s$ studies from each class to compute class prototypes, while a query set containing $N_q$ studies is used to classify new examples based on their distance from these prototypes in a learned embedding space. In our work, we employ a fundamental prototypical fine-tuning approach to adapt a medical text PLM for biomedical text mining while preserving essential domain-specific knowledge.

\subsection*{Data Acquisition and Preprocessing}
We retrieved 45,403 human studies from the GEO database up to April 2023. For each of the abstracts, we extracted titles, summaries, and experimental designs as input data for analysis. A proprietary pattern matching algorithm developed by Takeda Pharmaceutical identified ICI treatment studies using three sets of keywords: immunotherapy (for example, `PD-1', `CTLA-4', `checkpoint'), cancer (for example, `melanoma', `lung cancer', `breast cancer'), treatment (for example, `therapy,' `administration', `dose'). Due to intellectual property and legal considerations, the full details of the algorithm cannot be disclosed.

From 1,116 algorithm-flagged studies (any matched keyword), Takeda's computational biology team manually confirmed 121 positives (118 that corresponded to the three sets of keywords) and 1,105 negatives. The case review was conducted by a senior scientist.

\subsection*{Model Architecture}
PubMedBERT was used as the backbone to extract study-level embeddings, denoted by $\mathbf{h}_i$. Two approaches were compared: (1) the [CLS] token method, which uses the hidden state of the [CLS] token pre-trained for sentence-level tasks traditionally employed for BERT classification, and (2) mean pooling, which aggregates the contextualized representations of all tokens. Both methods were tuned during hyperparameter optimization to generate optimal study-level embeddings.

\begin{equation}
    \begin{gathered}
        \mathbf{h}_i= \begin{cases}\mathbf{h}_{[\text {CLS }]} & \text { (CLS strategy) } \\ \frac{1}{L} \sum_{j=1}^L \mathbf{h}_j & \text { (Mean pooling) }\end{cases}
    \end{gathered}
    \label{eq:Output_attention}
\end{equation}

Within each episode, support samples are randomly drawn from $C_{pos}$ and $C_{neg}$ to compute the class prototypes $P_{pos}$ and $P_{neg}$. respectively. The prototypes are computed as the mean of the support set embeddings:

\begin{equation}
    \begin{gathered}
        \mathbf{P}_k=\frac{1}{\left|\mathcal{S}_k\right|} \sum_{\left(x_i, y_i\right) \in \mathcal{S}_k} \mathbf{h}_i, \quad k \in\{pos, neg\}
    \end{gathered}
    \label{eq:Output_attention}
\end{equation}

To classify query studies, we evaluated two distance metrics between query embeddings $h_{i}$ and class prototypes $P_{k}$ in hyperparameter tuning. Specifically, we considered the (1) Euclidean distance, which measures the absolute differences between embeddings, and (2) cosine similarity, which captures the angular relationship between embeddings. Although the original prototypical network paper \cite{snell_prototypical_2017} demonstrated that the Euclidean distance generally outperformed cosine distance, a subsequent study of few-shot text classification with prototypical network favored cosine similarity{\cite{LIU2025111497}}. 

\begin{equation}
    \begin{gathered}
    d_{\mathrm{Euc}}\left(\mathbf{h}_i, \mathbf{P}_k\right)=\left\|\mathbf{h}_i-\mathbf{P}_k\right\|_2\\
        d_{\mathrm{Cos}}\left(\mathbf{h}_i, \mathbf{P}_k\right)=1-\frac{\mathbf{h}_i \cdot \mathbf{P}_k}{\left\|\mathbf{h}_i\right\|\left\|\mathbf{P}_k\right\|}
    \end{gathered}
    \label{eq:Output_attention}
\end{equation}

The probabilities of query samples belong to each class, prototypes are assigned via softmax over distances:

\begin{equation}
    \begin{gathered}
        P\left(y=k \mid x_i\right)=\frac{\exp \left(-d\left(\mathbf{h}_i, \mathbf{c}_k\right)\right)}{\sum_{k^{\prime}} \exp \left(-d\left(\mathbf{h}_i, \mathbf{c}_{k^{\prime}}\right)\right)}
    \end{gathered}
    \label{eq:Output_attention}
\end{equation}

 To preserve domain-specific information from medical texts, we integrated LoRA into the query and value matrices of all transformer layers in PubMedBERT, enabling parameter-efficient fine-tuning. This approach adds low-rank adaptors to the self-attention weight matrices $\left\{\boldsymbol{W^{\text{query}}}, \boldsymbol{W^{\text{value}}}\right\} \in \mathbb{R}^{d \times d}$ by computing $\boldsymbol{W} = \boldsymbol{W}_0 + \Delta \boldsymbol{W}$ , where $\Delta \boldsymbol{W}$ is decomposed into trainable matrices $\boldsymbol{A} \in \mathbb{R}^{d \times r}$, $\boldsymbol{B} \in \mathbb{R}^{r \times d}$ where $r \ll d$ ($\Delta \boldsymbol{W} = \boldsymbol{A} \boldsymbol{B}$). During training, the original PubMedBERT weights remain frozen, and only the LoRA parameters ($\boldsymbol{A},\boldsymbol{B}$) are updated. This reduces trainable parameters significantly, as each LoRA pair introduce only $2dr$ parameters per layer instead of $d^2$.


During training, the model minimizes cross-entropy loss over predictions of query sets $Q$:

\begin{equation}
    \begin{gathered}
        \mathcal{L}=-\frac{1}{|\boldsymbol{\mathcal{Q}}|} \sum_{\left(x_i, y_i\right) \in \boldsymbol{\mathcal{Q}}} \log P\left(y=y_i \mid x_i\right)
    \end{gathered}
    \label{eq:Output_attention}
\end{equation}

\section*{Experiments}

\subsection*{Data Sources and Pattern Matching Algorithm}\label{data-sources}

Among 1116 expert-reviewed samples, we first selected 20 positive and 20 negative studies for training. 10 positive and 10 negative holdout cases were designated for the construction of prototypes exclusively used for hyperparameter tuning and testing. A validation set for hyperparameter tuning was constructed with 20 positive and 200 negative studies, and an independent testing set comprised 71 positive and 765 negative studies. After testing, the model was be applied to an unlabeled dataset of 44,287 studies for real-world deployment. Performance of internal testing is measured using precision, recall and F1-score for $C_{pos}$.

\begin{table}[H]
	\caption{\textbf{Dataset Construction}}
	\label{tab:submission}
\begin{center}
\begin{tabular}{|l|l|l|l|}
\hline
\textbf{Subset }          &\textbf{Positive Studies}	&\textbf{Negative Studies}    &\textbf{Purpose}\\
\hline
Training                       &20	    &20                 &Model Training\\
\hline 
Prototype Set	              &10	    &10                &Prototype Construction\\
\hline
Validation                    &20       &200                 &Hyperparameter tuning\\
\hline
Test	                    &71     &765               &Internal Testing\\
\hline
Unlabeled Deployment	                    &-     &44,287               &Real-world application(unlabled)\\
\hline
\end{tabular}
\end{center}

\end{table}

\subsection*{Baselines and Implementation Details}
We compared our method (ProtoBERT-LoRA) against four baselines to evaluate its efficacy in low-resource immunotherapy study identification: (1) Takeda’s proprietary rule-based algorithm, which relies on keyword matching; (2)PubMedBERT (Direct Full FT), Direct full finetune PubMedBERT for classification; (3) PubMedBERT (Direct Full FT + Proto), a hybrid approach where PubMedBERT is first fully fine-tuned, and the embeddings of the prototypes and queries are computed post hoc from the finetuned model, using euclidean distance as classification metrics for prototypical inference; (4) Conventional machine learning (ML) classifiers (Random Forest, SVM, XGBoost) trained using PubMedBERT [CLS] token embeddings as features.

For reproducibility, all deep learning experiments were conducted in the following settings. We used the AdamW optimizer \cite{loshchilov2019decoupledweightdecayregularization} with a weight decay parameter  $\beta$ of 0.01. The batch size was set to 20 with balanced support/query splits (5 positive + 5 negative support samples and 5 positive + 5 negative query samples). Early stopping was implemented by monitoring the validation F1 score with a patience of 8 epochs, and training was run for a total of 101 epochs. A learning rate scheduler was used to reduce the learning rate when the monitored metric stopped improving. In the episodic experiments, each epoch consisted of 10 episodes. Episodic training was performed using 5-shot support and query sets per class.

During validation and testing, prototypes for each class are constructed using 10 positive and 10 negative samples, while the remaining samples from the validation and test sets serve as queries.

\subsection*{Hyperparameter Tuning}

We systematically tuned our hyperparameters by varying the LoRA rank [4, 8, 16, 32, 64] and dropouts [0, 0.1, 0.2], while also testing learning rates [1e-3, 8e-4, 6e-4, 2e-4, 1e-4, 5e-5], LoRA alpha is set to rank $\times 2$. Additionally, we evaluated two distance metrics: Euclidean and cosine, and two pooling strategies, the [CLS] token and mean pooling. The optimal configuration was found to be a learning rate of 6e-4, a LoRA rank of 32 with LoRA alpha set to 64, the [CLS] token for pooling, and the Euclidean distance metric, which yielded a maximum validation F1 score of 0.5. 

For our deep learning baseline methods and ablation studies, we only skip tuning the LoRA-specific parameters when LoRA is not applied, and similarly, we do not tune distance metrics and pooling strategies if the prototypical network is not used. All other components are tuned as needed. For conventional ML baseline models, Logistic Regression was tuned over C values [0.01, 0.1, 1, 10, 100] with an `l2' penalty and `lbfgs' solver; SVM was optimized by varying C over [0.1, 1, 10], exploring both `linear' and `rbf' kernels along with gamma settings of `scale' and `auto'; Random Forest was tuned for num estimators [50, 100, 200] and max depth [None, 10, 20]; and XGBoost was optimized with num estimators [50, 100, 200], max depth [3, 5, 7], and learning rate [0.01, 0.1, 0.2].

\section*{Results}
\subsection*{Performance Comparison}

ProtoBERT-LoRA achieves superior performance in immunotherapy study identification, attaining a 0.909 accuracy, 0.481 precision, and 0.624 F1-score, significantly outperforming all baselines. While the rule-based method achieves high recall (0.944), which is expected since the test set is curated by keyword matching, its poor precision (0.245) underscores limitations in resolving contextual ambiguities. ProtoBERT-LoRA balances precision and recall (0.481/0.887), surpassing the best conventional classifier (Logistic Regression) by \textbf{0.29\% }in F1 (0.624 vs. 0.483), demonstrating its efficacy in handling class imbalance and semantic nuance.

Conventional ML models (e.g., Random Forest, SVM) achieve moderate recall (0.704–0.802) but suffer from low precision (0.298–0.361), failing to generalize beyond surface-level keyword overlap. Directly fine-tuned PubMedBERT attains near-perfect recall (0.916) but severely compromises precision (0.259), likely overfitting to sparse training signals. Augmenting PubMedBERT with post-hoc prototypical inference marginally improves precision (0.285) and F1 (0.435) without recall gains, highlighting the necessity of jointly optimizing prototypes and embeddings during training. As shown in Figure 2, ProtoBERT-LoRA’s embeddings (c) forms a tighter cluster around the positive prototype region with clearer separation from the negatives, contrasting the diffuse distributions of vanilla pre-trained (a) and fine-tuned PubMedBERT (b).

\begin{table}[H]
\caption{\textbf{Performance Comparison on Immunotherapy Study Identification}}
\label{tab:submission}
\begin{center}
\begin{tabular}{|l|c|c|c|c|c|}
\hline
\textbf{Method} & \textbf{Accuracy} & \textbf{Precision} & \textbf{Recall} & \textbf{F1}& \textbf{Pred Positives (~44K unlabled)} \\
\hline
Rule-Based (Keyword Matching) & 0.749 & 0.245 & \textbf{0.944} & 0.390 &-\\
\hline
Random Forest & 0.847 & 0.333 & 0.802 & 0.471& 1461(\%3.3)\\
Logistic Regression & 0.867 & 0.361 & 0.732 & 0.483& 1115(\%2.5)\\
SVM & 0.843 & 0.319 & 0.747 & 0.447& 1293(\%2.9)\\
XGBoost & 0.834 & 0.298 & 0.704 & 0.418 & 1792(\%4.0)\\
\hline
PubMedBERT (Direct Full FT) & 0.770 & 0.259 & 0.916 & 0.404&2363(\%5.3) \\
PubMedBERT (Direct Full FT + Proto) & 0.798 & 0.285 & 0.916 & 0.435&1736(\%4.0) \\
\hline
\textbf{ProtoBERT-LoRA (Ours)} & \textbf{0.909} & \textbf{0.481} & 0.887 & \textbf{0.624} &\textbf{198(\%0.4)}\\
\hline
\end{tabular}
\end{center}
\end{table}

When applied to 44,287 unlabeled studies not flagged (i.e., no keywords matched) by the rule-based system, ProtoBERT-LoRA identified \textbf{198 (0.4\%)} potential studies for review. In comparison, PubMedBERT (direct fine-tuning) predicted \textbf{2,363 (5.3\%)}, PubMedBERT with post-hoc prototypes \textbf{1,736 (4.0\%)}, and conventional ML classifiers (Logistic Regression: \textbf{1,115 (2.5\%)}, SVM: \textbf{1,293 (2.9\%)}, Random Forest: \textbf{1,461 (3.3\%)}, XGBoost: \textbf{1,792 (4.0\%)}) produced substantially higher volumes. Manual review of ProtoBERT-LoRA’s 198 candidates revealed 5 additional relevant studies missed by the rule-based system, which lacked the predefined keywords, demonstrating the model’s ability to resolve contextual ambiguities and generalize beyond lexical patterns.

\begin{figure}[htbp]
    \centering
    \begin{subfigure}[b]{0.33\textwidth}
        \includegraphics[width=\linewidth]{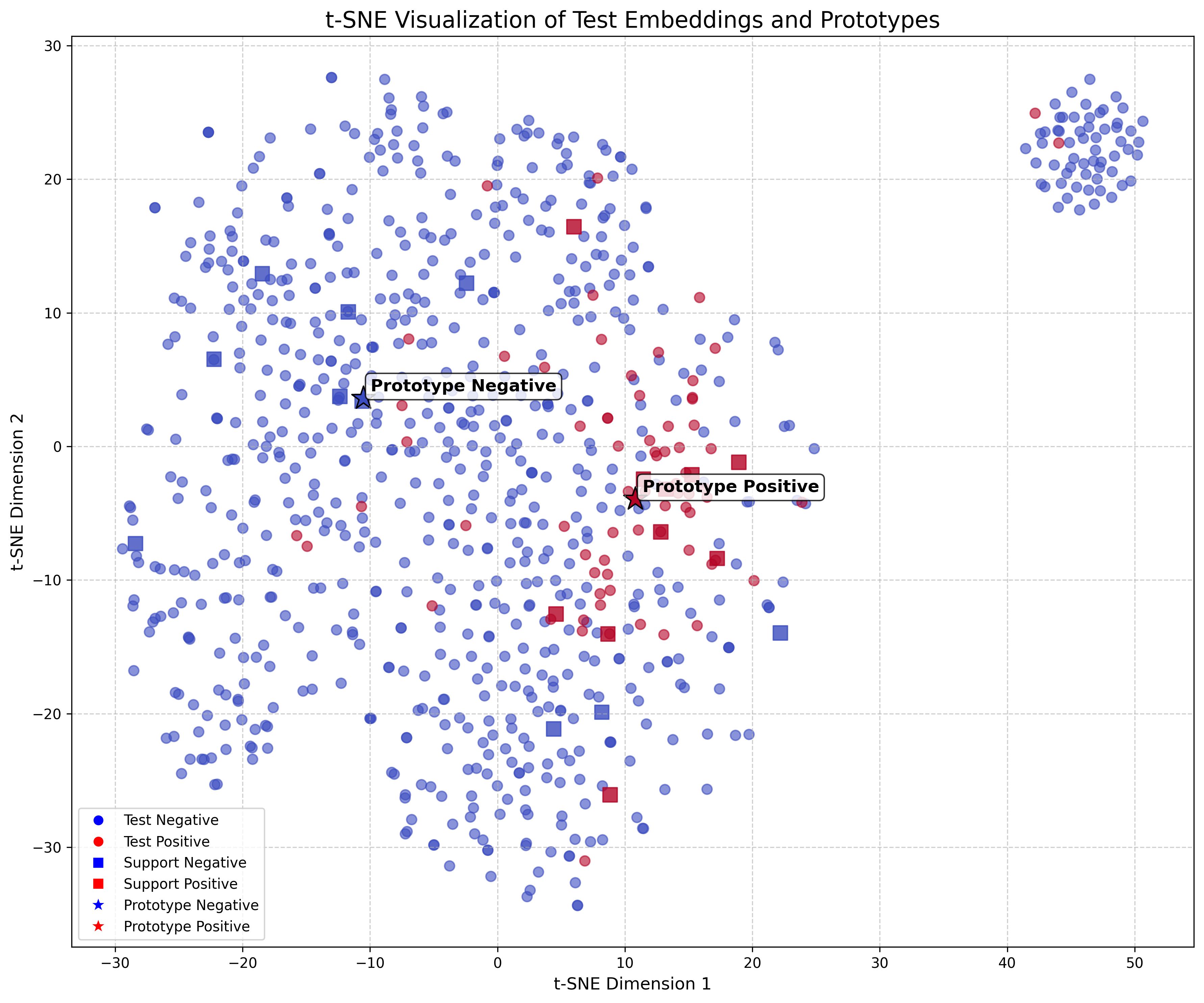}
        \caption{Vanilla PubMedBERT}
        \label{fig:a}
    \end{subfigure}
    \hfill
    \begin{subfigure}[b]{0.33\textwidth}
        \includegraphics[width=\linewidth]{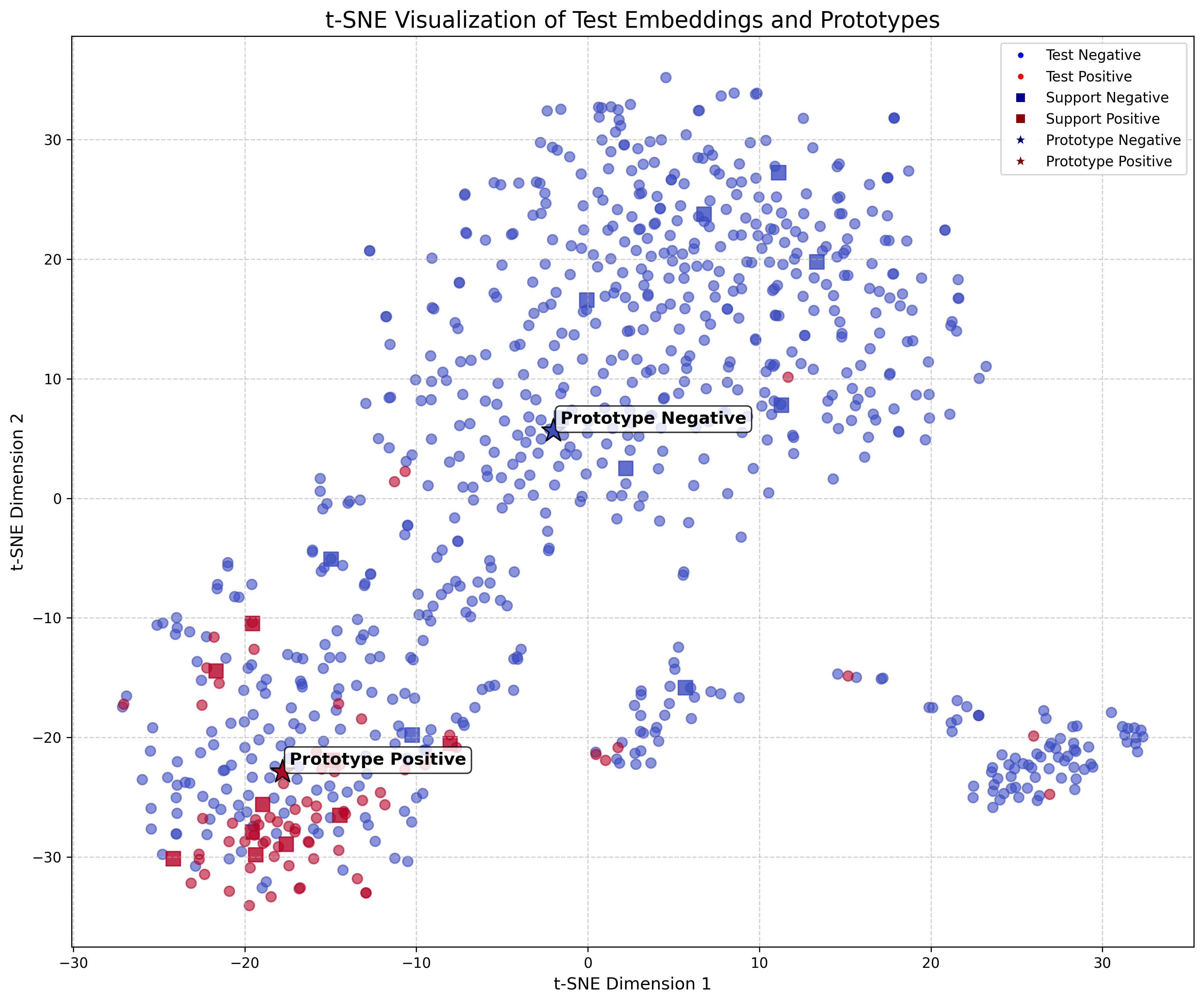}
        \caption{PubMedBERT (Direct Full FT)}
        \label{fig:b}
    \end{subfigure}
    \hfill
    \begin{subfigure}[b]{0.33\textwidth}
        \includegraphics[width=\linewidth]{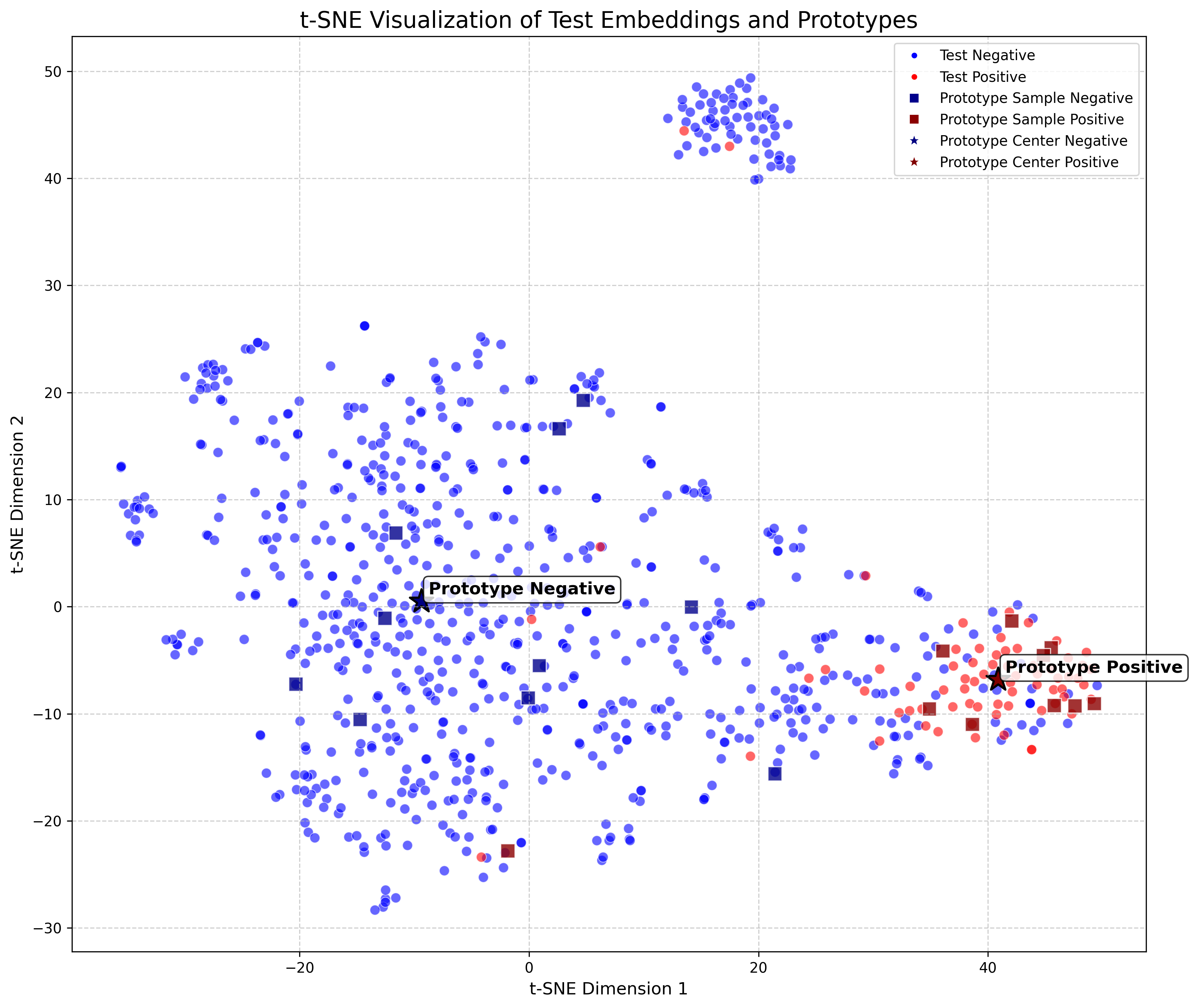}
        \caption{ProtoBERT-LoRA}
        \label{fig:c}
    \end{subfigure}
    \caption{t-SNE visualizations of test set embeddings from different model configurations. 
(a) Vanilla Pre-trained PubMedBERT without any task-specific adaptation. 
(b) PubMedBERT after conventional full fine-tuning. 
(c) ProtoBERT-LoRA, showing embeddings from the prototypical fine-tuning approach with LoRA adapters.}
    \label{fig:abc}
\end{figure}

\subsection*{Ablation Study}

In our ablation study, we evaluated the impact of the LoRA adapter and the prototypical learning framework by comparing three model variants: (1) vanilla PubMedBERT + Proto: Directly perform prototypical inference using the pretrained PubMedBERT by computing prototypes from the prototypical set and classifying via Euclidean distance. (2) PubMedBERT (LoRA FT) represents a model directly fine-tuned using a LoRA adapter; (3) PubMedBERT (LoRA FT + Proto) follows a hybrid approach where PubMedBERT is first fine-tuned with LoRA, and then the prototype and query embeddings are extracted from the fine-tuned model, with Euclidean distance used for classification; and (4) ProtoBERT without LoRA, which employs prototypical finetuning on PubMedBERT without incorporating LoRA. This comparison allows us to understand the individual contributions of the LoRA adapter and the prototypical learning mechanism to the overall performance.

\begin{table}[H]
\caption{\textbf{Ablation Study Comparison of Methods}}
\label{tab:submission}
\begin{center}
\begin{tabular}{|l|c|c|c|c|}
\hline
\textbf{Method} & \textbf{Accuracy} & \textbf{Precision} & \textbf{Recall} & \textbf{F1}\\
\hline
vanilla PubMedBERT + Proto & 0.645 & 0.164 & 0.775 & 0.270\\
PubMedBERT (LoRA FT) & 0.831 & 0.326 & \textbf{0.930} & 0.484\\
PubMedBERT (LoRA FT + Proto) & 0.871 & 0.389 & 0.916 & 0.546 \\
ProtoBERT (no LoRA) & 0.879 & 0.405 & 0.901 & 0.559\\
\hline
\textbf{ProtoBERT-LoRA (Ours)} & \textbf{0.909} & \textbf{0.481} & 0.887 & \textbf{0.624}\\
\hline
\end{tabular}
\end{center}
\end{table}

Among the four variants evaluated in the ablation study, vanilla PubMedBERT + Proto (with frozen embeddings) achieved poorest performance (F1: 0.270, recall: 0.775, precision: 0.164), highlighting the need for task-specific adaptation. PubMedBERT (LoRA FT) improved F1 to 0.484, although recall (0.930) remained much higher than precision (0.326). Adding post-hoc prototypical inference (LoRA FT + Proto) further increased F1 (0.546) and precision (0.389) with comparable recall (0.916). Employing prototypical finetuning without LoRA (ProtoBERT no LoRA) further improved precision (0.405) and F1 (0.559) while maintaining similar recall (0.916). Finally, with LoRA (ProtoBERT-LoRA), both precision (0.481) and F1 (0.624) improved greatly and reached the highest among variants, with only a slight trade-off in recall (0.887). Compared to PubMedBERT (LoRA FT), which scored the highest recall, our method improved accuracy by \textbf{0.078}, precision by \textbf{0.155}, and F1 by \textbf{0.14}, with a small reduction in recall of 0.043.

\section*{Discussion}
In this study, we addressed the challenge of identifying rare immunotherapy studies (ICI-exposed) in large genomic databases, a task often faced with challenges due to class imbalance, limited labeled data and sematic ambiguity. We proposed a hybrid approach, ProtoBERT-LoRA, that combined (i) a domain-specific transformer encoder (PubMedBERT), (ii) parameter-efficient fine-tuning via Low-Rank Adaptation (LoRA), and (iii) a prototypical network to enforce robust distance-based classification in low-resource settings.

Our results demonstrated that the proposed ProtoBERT-LoRA substantially outperformed all baseline methods in terms of both precision and F1 score for the ICI study classification. Traditional keyword- or rule-based approaches yielded high recall but suffered from very low precision. This may be due to their inability to interpret context, and matched keywords may not always indicate relevant immunotherapy studies. In contrast, direct fine-tuning of PubMedBERT achieved higher recall but still struggled with precision, implying an overfitting risk when labeled data are extremely limited. Even though applying prototypical inference post hoc to a full-finetuned encoder offered some improvement, our framework achieved superior performance by jointly optimizing prototypes and embeddings during training. 

By integrating prototypical learning directly into the fine-tuning process, our model learns to generate more coherent class representations (prototypes) in the latent space, enhancing intra-class compactness and inter-class separation. The random selection process during episodic training further mitigates overfitting by diversifying training instances. This framework achieves a balanced trade-off between precision (0.481) and recall (0.887), leading to the highest F1-score (0.624) among all evaluated methods. Ablation studies demonstrate that combining prototypical learning with LoRA yields the most significant performance improvements, emphasizing the synergistic effect of these two strategies.

When applied to $\sim$44,000 unlabeled studies lacking any immunotherapy keywords defined in our rule-based algorithm, ProtoBERT-LoRA successfully identified clinically relevant cases missed by the keyword-driven approach. This indicates that while the rule-based method relied on an exhaustive predefined keyword list to capture immunotherapy-related studies, ProtoBERT-LoRA generalized beyond its training data, detecting nuanced ``edge cases" where immunotherapy signals were implicitly expressed or contextually obscured. Compared to baseline algorithms (which flagged $>$1,000 cases for review), our approach reduced the manual review burden by 82\%--highlighting its precision in prioritizing high-probability candidates while filtering noise.

Our study has several limitations. First, the model’s reliance on keyword-preprocessed data introduces inherent biases: training and testing on studies filtered by predefined immunotherapy terms can limit exposure to edge cases or novel terminology, potentially reducing generalizability to unfiltered datasets. This dependency may also create a self-reinforcing cycle, which potentially constraining the model’s ability to detect immunotherapy concepts described outside the initial keyword lists (e.g., emerging therapies). Second, extreme class imbalance, with only 20 positive training examples, introduces overfitting risks, even with regularization via prototypical networks and LoRA. Such limited data may not adequately represent the heterogeneity of immunotherapy studies, particularly rare subtypes or evolving modalities. Third, while our model outperformed rule-based methods, we did not explicitly evaluate its semantic understanding of immunotherapy concepts. To address these limitations, future work could integrate active learning to iteratively refine keyword rules using high-confidence predictions, incorporate domain ontologies (e.g., MeSH terms) for standardized concept representation, or leverage semi-supervised learning to mitigate data scarcity. Adversarial testing\cite{He_2023} and human-in-the-loop validation could further evaluate semantic robustness, ensuring clinical interpretability.

Despite its limitations, this study presents a tool for cancer research to effectively recruit relevant study data in low-resource settings. ProtoBERT-LoRA could streamline the identification process, reducing manual review efforts, and allowing researchers to focus on high-quality studies, potentially accelerating biomarker discovery and therapeutic target identification in oncology with faster data curation. Moreover, its ability to learn class prototypes and fine-tune embeddings holds potential for adaptation to other low-resource tasks, such as identifying rare diseases or subtle phenotypic patterns, contributing broadly to other biomedical research and applications.

\subsection*{Conclusion}
Our study demonstrates that integrating prototypical learning with LoRA fine-tuning on PubMedBERT provides a robust framework for identifying rare immunotherapy studies from large genomic databases, achieving the highest accuracy (0.909), precision (0.481), and F1 score (0.664) with a comparable recall (0.887) relative to baseline algorithms and their ablation variants. The algorithm is able to identify cases that were completely missed by traditional pattern matching methods, capturing subtle ``edge cases" where immunotherapy signals are less obvious. This approach improves the precision of the study identification while reducing the need for manual review. It holds promise for extending biomedical text mining techniques to other low-resource domains, potentially accelerating biomarker discovery and therapeutic target identification. Future research should focus on expanding training datasets, incorporating advanced active learning strategies for generalization, and evaluating its semantic understanding in diverse biomedical contexts.
\subparagraph{Acknowledgments}
This study is an extension of an internship project completed during the summer of 2024 at the Takeda Data Science Institute, further developed by the Department of Biomedical Informatics at Johns Hopkins University. The views expressed in this work are solely those of the authors and do not necessarily reflect the official opinions or endorsements of Takeda or Johns Hopkins University.

\makeatletter
\renewcommand{\@biblabel}[1]{\hfill #1.}
\makeatother

\bibliographystyle{vancouver}
\bibliography{reference}  
\end{document}